\documentclass{article}

\usepackage{journal2e_v1}
\usepackage[a4paper]{geometry}

\usepackage{amssymb}
\usepackage{booktabs}
\usepackage{lmodern}
\usepackage{mathtools}
\usepackage{nth}
\usepackage{graphicx}
\usepackage{subfigure}

\makeindex

\def\A{{\bf A}}

\def\B{{\bf B}}
\def\b{{\bf b}}

\def\D{{\bf D}}

\def\L{{\bf L}}

\def\X{{\bf X}}

\def\T{{\mathcal T}}

\def\Z{{\bf Z}}
\def\M{{\bf M}}

\def\U{{\bf U}}

\def\V{{\bf V}}

\def\W{{\bf W}}

\def\0{{\bf 0}}
\def\1{{\bf 1}}
\def\wrt{{w.r.t.\ }}

\def\FB{{\mathbb F}}

\def\rank{\operatorname{rank}}

\def\vect{\operatorname{vec}}

\begin{document}


\title{Compression of Fully-Connected Layer in Neural Network by Kronecker
Product}

\author{
\name Shuchang Zhou     \\
\addr State Key Laboratory of Computer Architecture \\
Institute of Computing Technology \\
Chinese Academy of Sciences, Beijing, China\\
and Megvii Inc. \\
\texttt{shuchang.zhou@gmail.com} \\
          \AND
\name Jia-Nan Wu \\
 \addr Department of Computer Science \& Engineering \\
Shanghai Jiao Tong University \\
800 Dong Chuan Road, Shanghai, China 200240 \\
              \texttt{tankeco@sjtu.edu.cn}
}

\maketitle

\section{Abstract}
In this paper we propose and study a technique to reduce the number of
parameters and computation time in fully-connected layers of neural networks
using Kronecker product, at a mild cost of the prediction quality. The technique
proceeds by replacing Fully-Connected layers with so-called Kronecker
Fully-Connected layers, where the weight matrices of the FC layers are
approximated by linear combinations of multiple Kronecker products of smaller
matrices. In particular, given a model trained on SVHN dataset, we are able
to construct a new KFC model with 73\% reduction in total number of parameters,
while the error only rises mildly. In contrast, using low-rank method can only achieve
35\% reduction in total number of parameters given similar quality degradation
allowance.
If we only compare the KFC layer with its counterpart fully-connected layer, the
reduction in the number of parameters exceeds 99\%. The amount of
computation is also reduced as we replace matrix product of the large matrices
in FC layers with matrix products of a few smaller matrices in KFC layers.
Further experiments on MNIST, SVHN and some Chinese Character recognition models also demonstrate effectiveness of our technique.

\section{Introduction}
Model approximation aims at reducing the number of parameters and amount of
computation of neural network models, while keeping the quality of prediction
results mostly the same.\footnote{In some circumstances, as less number
of model parameters reduce the effect of overfitting, model approximation sometimes
leads to more accurate predictions.} Model approximation is important for real
world application of neural network to satisfy the time and storage constraints
of the applications.

In general, given a neural network $\tilde{f}(\cdot;\tilde{\theta})$, we want to
construct another neural network $f(\cdot;\theta)$ within some pre-specified
resource constraint, and minimize the differences between the outputs of two
functions on the possible inputs.
An example setup is to directly minimize the differences between the
output of the two functions:
\begin{align}
\label{align:model_approximation}
\inf_{\theta} \sum_i d(f(x_i;\theta), \tilde{f}(x_i;\tilde{\theta}))
\text{,}
\end{align}
where $d$ is some distance function and $x_i$ runs over all input data.

The formulation \ref{align:model_approximation} does not give any constraints
between the structure of $f$ and $\tilde{f}$, meaning that any model can be used
to approximate another model. In practice, a structural similar model is often
used to approximate another model. In this case, model approximation may be
approached in a modular fashion \wrt to each layer.

\subsection{Low Rank Model Approximation}

Low rank approximation in linear regression dates back to
\cite{anderson1951estimating}. In \cite{sainath2013low, liao2013large,
xue2013restructuring, zhang2014extracting, denton2014exploiting}, low rank
approximation of fully-connected layer is used; and
\cite{jaderberg2014speeding, rigamonti2013learning, DBLP:journals/corr/LebedevGROL14} considered low rank
approximation of convolution layer. \cite{zhang2014efficient} considered
approximation of multiple layers with nonlinear activations.

We first outline the low rank approximation method below. The fully-connected
layer widely used in neural network construction may be formulated as:
\begin{align}
\L_{a} = h(\L_{a-1} \M_a + \b_a) \text{,}
\end{align}

where $\L_i$ is the output of the $i$-th layer of the neural network,
$\M_a$ is often referred to as ``weight term'' and $\b_a$ as ``bias term'' of
the $a$-th layer.


As the coefficients of the weight term in the fully-connected layers are
organized into matrices, it is possible to perform low-rank approximation of
these matrices to achieve an approximation of the layer, and consequently the
whole model.
Given Singular Value Decomposition of a matrix $\M = \U \D \V^*$, where $\U,
\V$ are unitary matrices and $\D$ is a diagonal matrix with the diagonal made
up of singular values of $\M$, a rank-$k$ approximation of $\M\in\FB^{m\times
n}$ is:
\begin{align}
\M \approx \M_k = \tilde{\U} \tilde{\D} \tilde{\V}^*\text{ , where }
\tilde{\U}\in\FB^{m\times k}\text{, }\tilde{\D}\in\FB^{k\times
k}\text{, }\tilde{\V}\in\FB^{n\times k}
\text{,}
\end{align}

where $\tilde{\U}$ and $\tilde{\V}$ are the first $k$-columns of the $\U$ and
$\V$ respectively, and $\tilde{\D}$ is a diagonal matrix made up of the largest
$k$ entries of $\D$.

In this case approximation by SVD is optimal in the sense that the following
holds \cite{horn1991topics}:
\begin{align}
\M_r = \inf_{\X} \|\X - \M\|_F \text{ s.t. } \rank(\X) \le r \text{,}
\end{align}
and
\begin{align}
\M_r = \inf_{\X} \|\X - \M\|_2 \text{ s.t. } \rank(\X) \le r \text{.}
\end{align}

The approximate fully connected layer induced by SVD is:
\begin{align}
\Z & = \L_{a-1} \tilde{\U} \\
\L_{a} & = h(\Z \D \V^* + \b_a) \text{,}
\end{align}

In the modular representation of neural network, this means that the original
fully connected layer is now replaced by two consequent fully-connected layers.


%
%

However, the above post-processing approach only ensures getting an optimal
approximation of $\M$ under the rank constraint, while there is still no
guarantee that such an approximation is optimal \wrt the input data.
I.e., the optimum of the following may well be different from the rank-$r$
approximation $\M_r$ \wrt some given input $\X$:
\begin{align}
\inf_{\Z} \|\Z\X - \M\X\|_F \text{ s.t. } \rank(\Z) \le r \text{.}
\end{align}

Hence it is often necessary for the resulting low-rank model
$\tilde{\mathcal{M}}$ to be trained for a few more epochs on the input,
which is also known as the ``fine-tuning'' process.

Alternatively, we note that the rank constraint can be enforced by the
following structural requirement for $\X\in\FB^{m\times n}$:
\begin{align}
\rank(\X) \le r \Leftrightarrow \exists \A\in\FB^{m\times r},\,
\B\in\FB^{r\times n}\text{ s.t. } \X = \A \B\text{.}
\end{align}

In light of this, if we want to impose a rank constraint on a fully-connected
layer $L(\M, g)$ in a neural network where $\M\in\FB^{m\times n}$, we can
replace that layer with two consecutive layers $L_1(\B, g_1)$ and $L_2(\A,
g_2)$, where $g_1(x) = x$, $g_2 = g$, and $\M = \A \B$ where $\A\in\FB^{m\times
r},\,\B\in\FB^{r\times n}$, and then train the structurally constrained neural
network on the training data.



As a third method, a regularization term inducing low rank matrices may be
imposed on the weight matrices. In this case, the training of a $k$-layer model is modified
to be:
\begin{align}
\inf_{\theta} \sum_i f(x_i;\theta) + \sum_{j=1}^k r_j(\theta)
\text{,}
\end{align}

where $r$ is the regularization term. For the weight term of the FC layers,
conceptually we may use the matrix rank function as the regularization term. However, as the rank function is only well-defined for infinite-precision
numbers, nuclear norm may be used as its convex proxy
\cite{jaderberg2014speeding, Recht:2010:GMS:1958515.1958520}.

\section{Model Approximation by Kronecker Product}
Next we propose to use Kronecker product of matrices of particular shapes for
model approximation in Section~\ref{subsec:kronecker-product}. We also outline
the relationship between the Kronecker product approximation and low-rank
approximation in Section~\ref{subsec:rel-kp-lowrank}.

Below we measure the reduction in amount of computation by number of floating
point operations. In particular, we will assume the computation complexity of
two matrices of dimensions $M\times K$ and $K\times N$ to be $O(M K N)$, as many
neural network implementations \cite{Bastien-Theano-2012,
bergstra+al:2010-scipy, jia2014caffe, collobert2011torch7} have not used
algorithms of lower computation complexity for the typical inputs of the neural
networks. Our analysis is mostly immune to the ``hidden constant'' problem in
computation complexity analysis as the underlying computations of the
transformed model may also be carried out by matrix products.

\subsection{Weight Matrix Approximation by Kronecker Product}
\label{subsec:kronecker-product}

We next discuss how to use Kronecker product to approximate weight matrices of
FC layers, leading to construction of a new kind of layer which we call
Kronecker Fully-Connected layer. The idea originates from the observation that
for a matrix $\M \in \FB^{m\times n}$ where the dimensions are not prime
\footnote{In case any of $m$ and $n$ is prime, it is possible to add some extra dummy
feature or output class to make the dimensions dividable.}, we have
approximations like:
\begin{align}
\M = \M_1 \otimes \M_2 \text{,}
\end{align}
where $m = m_1 m_2$, $n = n_1 n_2$, $\M_1\in \FB^{m_1\times n_1}$, $\M_2\in
\FB^{m_2\times n_2}$.

Any factors of $m$ and $n$ may be selected as $m_1$ and $n_1$ in the above
formulation. However, in a Convolutional Neural Network, the input to a FC layer
may be a tensor of order 4, which has some natural shape constraints that we will try
to leverage in \ref{subsubsec:kp-approx-fc-4d-tensor}. Otherwise, when the input is
a matrix, we do not have natural choices of $m_1$ and $n_1$. We will explore
heuristics to pick $m_1$ and $n_1$ in \ref{subsubsec:kp-approx-fc-matrix}.

\subsubsection{Kronecker product approximation for fully-connected
layer with 4D tensor input}
\label{subsubsec:kp-approx-fc-4d-tensor}
In a convolutional layer processing images, the input data $\L_{a-1}$ may be
a tensor of order 4 as $\T_{nchw}$ where $n=1,2,\cdots,N$ runs over $N$
different instances of data, $c=1,2,\cdots,C$ runs over $C$ channels of the
given images, $h=1,2,\cdots,H$ runs over $H$ rows of the images, and
$w=1,2,\cdots,W$ runs over $W$ columns of the images. $\T$ is often reshaped
into a matrix before being fed into a fully connected layer as $\D_{nj}$, where
$n=1,2,\cdots,N$ runs over the $N$ different instances of data and
$j=1,2,\cdots,CHW$ runs over the combined dimension of channel, height, and
width of images. The weights of the fully-connected layer would then be a matrix
$\M_{jk}$ where $j=1,2,\cdots,CHW$ and $k=1,2,\cdots,K$ runs over output number
of channels. I.e., the layer may be written as:
\begin{align}
\D & = \text{Reshape}(\L_{a-1}) \\
\L_a & = h(\D \M + \b)
\text{.}
\end{align}

Though the reshaping transformation from $\T$ to $\D$ does not incur any loss in
pixel values of data, we note that the dimension information of the tensor of
order 4 is lost in the matrix representation. As a consequence, $\M$ has $CHWK$ number
of parameters.

Due to the shape of $\M$, we may propose a few kinds of structural constraint on
$\M$ by requiring $\M$ to be Kronecker product of matrices of particular shapes.
\subsubsection{Formulation I}
In this formulation, we require $\M = \M_1 \otimes \M_2 \otimes \M_3$, where
$\M_1\in\FB^{C\times K_1}, \M_2 \in \FB^{H\times K_2}, \M_3 \in \FB^{W\times
K_3}$, and $K=K_1 K_2 K_3$.
The number of parameters is reduced to $CK_1 + HK_2 + WK_3$. The underlying
assumption for this model is that the transformation is invariant across rows and
columns of the images.

\subsubsection{Formulation II}
In this formulation, we require $\M = \M_1 \otimes \M_2$, where
$\M_1\in\FB^{C\times K_1}, \M_2 \in \FB^{HW\times K_2}$, and $K = K_1 K_2$.
The number of parameters is reduced to $CK_1 + HWK_2$. The underlying assumption
for this model is that the channel transformation should be decoupled from the
spatial transformation.

\subsubsection{Formulation III}
In this formulation, we require $\M = \M_1 \otimes \M_2$, where
$\M_1\in\FB^{CH\times K_1}, \M_2 \in \FB^{W\times K_2}$, and $K = K_1 K_2$.
The number of parameters is reduced to $CHK_1 + WK_2$. The underlying
assumption for this model is that the transformation \wrt columns may be
decoupled.

\subsubsection{Formulation IV}
In this formulation, we require $\M = \M_1 \otimes \M_2$, where
$\M_1\in\FB^{CW\times K_1}, \M_2 \in \FB^{H\times K_2}$, and $K = K_1 K_2$.
The number of parameters is reduced to $CWK_1 + HK_2$. The underlying
assumption for this model is that the transformation \wrt rows may be
decoupled.

\subsubsection{Combined Formulations}
Note that the above four formulations may be linearly combined to produce more
possible kinds of formulations. It would be a design choice with respect to
trade off between the number of parameters, amount of computation and the
particular formulation to select.

\subsubsection{Kronecker product approximation for matrix input}
\label{subsubsec:kp-approx-fc-matrix}
For fully-connected layer whose input are matrices, there does not exist natural
dimensions to adopt for the shape of smaller weight matrices in KFC. Through
experiments, we find it possible to arbitrarily pick a decomposition of input
matrix dimensions to enforce the Kronecker product structural constraint. We
will refer to this formulation as KFCM.

Concretely, when input to a fully-connected layer is $\X \in \FB^{N\times C}$
and the weight matrix of the layer is $\W\in \FB^{C\times K}$, we can construct
approximation of $\W$ as:
\begin{align}
\tilde{\W} = \W_1 \otimes \W_2 \approx \W \text{,} 
\end{align}
where $C = C_1 C_2$, $K=K_1 K_2$, $\W_1 \in \FB^{C_1\times K_1}$ and $\W_2 \in
\FB^{C_2\times K_2}$.

The computation complexity will be reduced from $O(N C K)$ to $O(N C K
(\frac{1}{K_2} + \frac{1}{C_1})) = O(N C_2 C_1 K_1 + N C_2 K_1 K_2)$,
while the number of parameters will be reduced from $C K$ to $C_1 K_1 + C_2
K_2$.

Through experiments, we have found it sensible to pick $C_1 \approx \sqrt{C}$
and $K_1 \approx \sqrt{K}$.

As the choice of $C_1$ and $K_1$ above is arbitrary, we may use linear
combination of Kronecker products if matrices of different shapes for
approximation.

\begin{align}
\label{align:kp-multiple-shape}
\tilde{\W} = \sum_{j=1}^{J} \W_{1j} \otimes \W_{2j} \approx \W \text{,} 
\end{align}
where $\W_{1j} \in \FB^{C_{1j}\times K_{1j}}$ and $\W_{2j} \in
\FB^{C_{2j}\times K_{2j}}$.

\subsection{Relationship between Kronecker Product Constraint and Low Rank
Constraint}
\label{subsec:rel-kp-lowrank}
It turns out that factorization by Kronecker product is closely related to the
low rank approximation method. In fact, approximating a matrix $\M$ with
Kronecker product $\M_1 \otimes \M_2$ of two matrices may be casted into a Nearest Kronecker product Problem:
\begin{align}
\inf_{\M_1, \M_2} \|\M - \M_1 \otimes \M_2\|_F \text{.}
\end{align}

An equivalence relation in the above problem is given in
\cite{van1993approximation, van2000ubiquitous} as:
\begin{align}
\label{align:nkp-rank1}
\arg\inf_{\M_1, \M_2} \|\M - \M_1 \otimes \M_2\|_F =
\arg \inf_{\M_1, \M_2} \|\mathcal{R}(\M) - \vect{\M_1} (\vect{\M_2})^{\top}\|_F
\text{,}
\end{align}
where $\mathcal{R}(\M)$ is a matrix formed by a fixed reordering of entries
$\M$.

Note the right-hand side of formula~\ref{align:nkp-rank1} is a rank-1
approximation of matrix $\mathcal{R}(\M)$, hence has a closed form solution.
However, the above approximation is only optimal \wrt the parameters of the
weight matrices, but not \wrt the prediction quality over input data.

Similarly, though there are iterative algorithms for rank-1 approximation of
tensor \cite{friedland2013best, Lathauwer:2000:BRR:354353.354405}, the
optimality of the approximation is lost once input data distribution is taken
into consideration.

Hence in practice, we only use the Kronecker Product constraint to construct KFC
layers and optimize the values of the weights through the training process on
the input data.

\subsection{Extension to Sum of Kronecker Product}
Just as low-rank approximation may be extended beyond rank-1 to arbitrary number
of ranks, one could extend the Kronecker Product approximation to Sum of
Kronecker Product approximation. Concretely, one not the following
decomposition of $\M$:
\begin{align}
\M = \sum_{i=1}^{\rank(\mathcal{R}(\M))} \A_i \otimes \B_i
\text{.}
\end{align}

Hence it is possible to find $k$-approximations:
\begin{align}
\M \approx \sum_{i=1}^{k} \A_i \otimes \B_i
\text{.}
\end{align}

We can then generalize Formulation~I-IV in \ref{subsec:kronecker-product} to the
case of sum of Kronecker Product.

We may further combine the multiple shape formulation of
\ref{align:kp-multiple-shape} to get the general form of KFC layer:
\begin{align}
\label{align:kp-multiple-shape-multiple-sum}
\M \approx \sum_{j=1}^{J} \sum_{i=1}^{k} \A_{ij} \otimes \B_{ij}
\text{.}
\end{align}
where $\A_{ij} \in \FB^{C_{1j}\times K_{1j}}$ and $\B_{ij} \in
\FB^{C_{2j}\times K_{2j}}$.

\section{Empirical Evaluation of Kronecker product method}
We next empirically study the properties and efficacy of the Kronecker product method
and compare it with some other common low rank model approximation methods.

To make a fair comparison, for each dataset, we train a covolutional neural network with a fully-connected layer
as a baseline. Then we replace the fully-connected layer with different layers according to different methods and train
the new network until quality metrics stabilizes. We then compare KFC method
with low-rank method and the baseline model in terms of number of parameters and
prediction quality. We do the experiments based on implementation of KFC layers
in Theano\cite{bergstra+al:2010-scipy, Bastien-Theano-2012} framework.

As the running time may depend on particular implementation details of the KFC
and the Theano work, we do not report running time below. However, there is no
noticeable slow down in our experiments and the complexity analysis suggests
that there should be significant reduction in amount of computation.

\subsection{MNIST}
The MNIST dataset\cite{lecun1998gradient} consists of $28\times28$ grey scale images 
of handwritten digits. There are 60000 training images and 10000 test images. We 
select the last 10000 training images as validation set.

Our baseline model has $8$ layers and the first $6$ layers consist of four
convolutional layers and two pooling layers.
The 7th layer is the fully-connected layer and the 8th is the softmax output.
The input of the fully-connected layer is of size $32\times3\times3$, where $32$ is the number of channel and $3$ is the side length of
image patches(the mini-batch size is omitted). The output of the fully-connected layer is of size $256$, so the weight matrix
is of size $288\times256$.

CNN training is done with Adam\cite{kingma2014adam} with weight decay of 0.0001. Dropout\cite{hinton2012improving} of 0.5 is used 
on the fully-connected layer and KFC layer. $y=|\tanh{x}|$ is used as activation 
function. Initial learning rate is $1e-4$ for Adam.

Test results are listed in Table~\ref{tab:mnist_approximation}. The number of layer parameters means the number of parameters of the fully-connected 
layer or its counterpart layer(s). The number of model parameters is the number of the parameters of the whole model. The test error is the min-validation
model's test error.

In Cut-96 method, we use 96 output neurons instead of 256 in fully-connected layer. In the LowRank-96 method, we replace the fully-connected layer with two fully-connected layer where the first FC layer output size is 96 and the second FC layer output size is 256. In the KFC-II method, we replace the fully-connected layer with KFC layer using
formulation II with $K_1=64$ and $K_2=4$.  In the KFC-Combined method, we replace the fully-connected layer with KFC layer and linear combined the formulation II, III and IV($K_1=64, K_2=4$ in formulation II, $K_1=128, K_2=2$ in formulation III and IV).
\begin{table}[!ht] \centering \small
\caption{Comparison of using Low-Rank method and using KFC layers on MNIST
dataset}
\begin{center}
\begin{tabular}{p{2cm} p{5cm} p{5cm} p{1.5cm}}
    \toprule Methods  & \# of Layer Params(\%Reduction) & \# of Model Params(\%Reduction) & Test Error \\
    \midrule Baseline & 74.0K & 99.5K & 0.51\% \\
    \midrule Cut-96 & 27.8K(62.5\%) & 51.7K(48.1\%) & 0.58\% \\
    \midrule LowRank-96 & 52.6K(39.0\%) & 78.1K(21.6\%) & 0.54\% \\
    \midrule KFC-II &  2.1K(97.2\%)  & 27.7K(72.2\%) & 0.76\%\\
    \midrule KFC-Combined &  27.0K(63.51\%)  & 52.5K(47.2\%) & 0.57\%\\
\hline
\end{tabular}
\end{center} \label{tab:mnist_approximation}
\end{table}

\subsection{Street View House Numbers}
The SVHN dataset\cite{netzer2011reading} is a real-world digit recognition dataset 
consisting of photos of house numbers in Google Street View images. The
dataset comes in two formats and we consider the second format: 32-by-32 colored images centered 
around a single character. There are 73257 digits for training, 26032 digits 
for testing, and 531131 less difficult samples which can be used as extra training 
data. To build a validation set, we randomly select 400 images per class from 
training set and 200 images per class from extra training set 
as \cite{sermanet2012convolutional, goodfellow2013maxout} did.

Here we use a similar but larger neural network as used in MNIST to be the baseline. The input of the fully-connected layer
is of size $256\times5\times5$. The fully-connected layer has $256$ output neurons. Other implementation details are not changed.
Test results are listed in Table~\ref{tab:svhn_approximation}. In the Cut-$N$ method, we use $N$ output neurons instead of 256 in fully-connected layer. In the LowRank-$N$ method, we replace the fully-connected layer with two fully-connected layer where the first FC layer output size is $N$ and the second FC layer output size is 256. In the KFC-II method, we replace the fully-connected layer with KFC layer using formulation II with $K_1=64$ and $K_2=4$.  In the KFC-Combined method, we replace the fully-connected layer with KFC layer and linear combined the formulation II, III and IV($K_1=64, K_2=4$ in formulation II,  $K_1=128, K_2=2$ in formulation III and IV). In the KFC-Rank$N$ method, we use KFC formulation II with $K_1=64,K_2=2$ and extend it to rank $N$ with as described above.
\begin{table}[!ht] \centering \small
\caption{Comparison of using Low-Rank method and using KFC layers on SVHN dataset}
\begin{center}
\begin{tabular}{p{2cm} p{5cm} p{5cm} p{1.5cm}}
    \toprule Methods  & \# of Layer Params(\%Reduction) & \# of Model Params(\%Reduction) & Test Error \\
    \midrule Baseline & 1.64M & 2.20M & 2.57\% \\
    \midrule Cut-128 & 0.82M(50.0\%) & 1.38M(37.3\%) & 2.79\% \\
    \midrule Cut-64 & 0.41(25.0\%) & 0.97(55.9\%) & 3.19\% \\
    \midrule LowRank-128 & 0.85M(48.2\%) & 1.42M(35.7\%) & 3.02\% \\
    \midrule LowRank-64 & 0.43M(73.7\%) & 0.99M(55.1\%) & 3.67\% \\
    \midrule KFC-II & 0.016M(99.0\%) & 0.58M(73.7\%) & 3.33\% \\
    \midrule KFC-Combined & 0.34M(79.3\%) & 0.91M(58.6\%) & 2.60\% \\
    \midrule KFC-Rank10 & 0.17M(89.6\%) & 0.73M(66.8\%) & 3.19\% \\
\hline
\end{tabular}
\end{center} \label{tab:svhn_approximation}
\end{table}

\subsection{Chinese Character Recognition}
We also evaluate application of KFC to a Chinese character recognition model.
Our experiments are done on a private dataset for the moment and may extend to
other established Chinese character recognition datasets like
HCL2000(\cite{zhang2009hcl2000}) and CASIA-HWDB(\cite{liu2013online}).

For this task we also use a convolutional neural network. The distinguishing
feature of the neural network is that following the convolution and pooling
layers, it has two FC layers, one with 1536 hidden size, and the other with more
than 6000 hidden size.

The two FC layers happen to be different type. The 1st FC layer accepts tensor
as input and the 2nd FC layer accepts matrix as input. We apply KFC-I
formulation to 1st FC and KFCM to 2nd FC.

\begin{table}[!ht] \centering \small
\caption{Effect of using KFC layers on a Chinese
recognition dataset}
\begin{center}
\begin{tabular}{p{2cm} p{2cm} p{2cm} p{2cm} p{1.5cm}}
    \toprule Methods  & \%Reduction of 1st FC Layer Params & \%Reduction of 2nd
    FC Layer Params & \%Reduction of Total Params & Test Error \\
    \midrule Baseline & 0\% & 0\% & 0\% & 10.6\% \\
    \midrule KFC-II & 99.3\% & 0\% & 36.0\% & 11.6\% \\
    \midrule KFC-KFCM-rank1 & 98.7\% & 99.9\% & 94.5\% & 21.8\% \\
    \midrule KFC-KFCM-rank-10 & 93.3\% & 99.1\% & 91.8\% & 13.0\% \\
\hline
\end{tabular}
\end{center} \label{tab:svhn_approximation}
\end{table}

It can be seen KFC can significantly reduce the number of parameters. However,
in case of ``KFC and KFCM (rank=1)'', this also leads to serious degradation of
prediction quality. However, by increasing the rank from 1 to 10, we are able to
recover most of the lost prediction quality. Nevertheless, the rank-10 model is
still very small compared to the baseline model.

\section{Conclusion and Future Work}
In this paper, we propose and study methods for approximating the weight
matrices of fully-connected layers with sums of Kronecker product of smaller
matrices, resulting in a new type of layer which we call Kronecker
Fully-Connected layer.
We consider both the cases when input to the fully-connected layer is a tensor
of order 4 and when the input is a matrix. We have found that using the KFC
layer can significantly reduce the number of parameters and amount of
computation in experiments on MNIST, SVHN and Chinese character recognition.

As future work, we note that when weight parameters of a convolutional layer is
a tensor of order 4 as $\mathcal{T}\in \FB^{K\times C\times H\times W}$, it can
be represented as a collection of $H\times W$ matrices $\mathbf{T}_{hw}$. We
can then approximate each matrix by Kronecker products as $\mathbf{T}_{hw} =
\A_{hw}\otimes\B_{hw}$ following KFCM formulation, and apply the other
techniques outlined in this paper. It is also noted that the Kronecker product
technique may also be applied to other neural network architectures like
Recurrent Neural Network, for example approximating transition matrices with
linear combination of Kronecker products.

\bibliographystyle{spmpsci}
\bibliography{thesis}

\begin{thebibliography}{28}
\providecommand{\natexlab}[1]{#1}
\providecommand{\url}[1]{\texttt{#1}}
\expandafter\ifx\csname urlstyle\endcsname\relax
  \providecommand{\doi}[1]{doi: #1}\else
  \providecommand{\doi}{doi: \begingroup \urlstyle{rm}\Url}\fi

\bibitem[Anderson(1951)]{anderson1951estimating}
Theodore~Wilbur Anderson.
\newblock Estimating linear restrictions on regression coefficients for
  multivariate normal distributions.
\newblock \emph{The Annals of Mathematical Statistics}, pages 327--351, 1951.

\bibitem[Bastien et~al.(2012)Bastien, Lamblin, Pascanu, Bergstra, Goodfellow,
  Bergeron, Bouchard, and Bengio]{Bastien-Theano-2012}
Fr{\'{e}}d{\'{e}}ric Bastien, Pascal Lamblin, Razvan Pascanu, James Bergstra,
  Ian~J. Goodfellow, Arnaud Bergeron, Nicolas Bouchard, and Yoshua Bengio.
\newblock Theano: new features and speed improvements.
\newblock Deep Learning and Unsupervised Feature Learning NIPS 2012 Workshop,
  2012.

\bibitem[Bergstra et~al.(2010)Bergstra, Breuleux, Bastien, Lamblin, Pascanu,
  Desjardins, Turian, Warde-Farley, and Bengio]{bergstra+al:2010-scipy}
James Bergstra, Olivier Breuleux, Fr{\'{e}}d{\'{e}}ric Bastien, Pascal Lamblin,
  Razvan Pascanu, Guillaume Desjardins, Joseph Turian, David Warde-Farley, and
  Yoshua Bengio.
\newblock Theano: a {CPU} and {GPU} math expression compiler.
\newblock In \emph{Proceedings of the Python for Scientific Computing
  Conference ({SciPy})}, June 2010.
\newblock Oral Presentation.

\bibitem[Collobert et~al.(2011)Collobert, Kavukcuoglu, and
  Farabet]{collobert2011torch7}
Ronan Collobert, Koray Kavukcuoglu, and Cl{\'e}ment Farabet.
\newblock Torch7: A matlab-like environment for machine learning.
\newblock In \emph{BigLearn, NIPS Workshop}, number EPFL-CONF-192376, 2011.

\bibitem[Denton et~al.(2014)Denton, Zaremba, Bruna, LeCun, and
  Fergus]{denton2014exploiting}
Emily~L Denton, Wojciech Zaremba, Joan Bruna, Yann LeCun, and Rob Fergus.
\newblock Exploiting linear structure within convolutional networks for
  efficient evaluation.
\newblock In \emph{Advances in Neural Information Processing Systems}, pages
  1269--1277, 2014.

\bibitem[Friedland et~al.(2013)Friedland, Mehrmann, Pajarola, and
  Suter]{friedland2013best}
Shmuel Friedland, Volker Mehrmann, Renato Pajarola, and SK~Suter.
\newblock On best rank one approximation of tensors.
\newblock \emph{Numerical Linear Algebra with Applications}, 20\penalty0
  (6):\penalty0 942--955, 2013.

\bibitem[Goodfellow et~al.(2013)Goodfellow, Warde-Farley, Mirza, Courville, and
  Bengio]{goodfellow2013maxout}
Ian~J Goodfellow, David Warde-Farley, Mehdi Mirza, Aaron Courville, and Yoshua
  Bengio.
\newblock Maxout networks.
\newblock \emph{arXiv preprint arXiv:1302.4389}, 2013.

\bibitem[Hinton et~al.(2012)Hinton, Srivastava, Krizhevsky, Sutskever, and
  Salakhutdinov]{hinton2012improving}
Geoffrey~E Hinton, Nitish Srivastava, Alex Krizhevsky, Ilya Sutskever, and
  Ruslan~R Salakhutdinov.
\newblock Improving neural networks by preventing co-adaptation of feature
  detectors.
\newblock \emph{arXiv preprint arXiv:1207.0580}, 2012.

\bibitem[Horn and Johnson(1991)]{horn1991topics}
Roger~A Horn and Charles~R Johnson.
\newblock \emph{Topics in matrix analysis}.
\newblock Cambridge university press, 1991.

\bibitem[Jaderberg et~al.(2014)Jaderberg, Vedaldi, and
  Zisserman]{jaderberg2014speeding}
Max Jaderberg, Andrea Vedaldi, and Andrew Zisserman.
\newblock Speeding up convolutional neural networks with low rank expansions.
\newblock \emph{arXiv preprint arXiv:1405.3866}, 2014.

\bibitem[Jia et~al.(2014)Jia, Shelhamer, Donahue, Karayev, Long, Girshick,
  Guadarrama, and Darrell]{jia2014caffe}
Yangqing Jia, Evan Shelhamer, Jeff Donahue, Sergey Karayev, Jonathan Long, Ross
  Girshick, Sergio Guadarrama, and Trevor Darrell.
\newblock Caffe: Convolutional architecture for fast feature embedding.
\newblock In \emph{Proceedings of the ACM International Conference on
  Multimedia}, pages 675--678. ACM, 2014.

\bibitem[Kingma and Ba(2014)]{kingma2014adam}
Diederik Kingma and Jimmy Ba.
\newblock Adam: A method for stochastic optimization.
\newblock \emph{arXiv preprint arXiv:1412.6980}, 2014.

\bibitem[Lathauwer et~al.(2000)Lathauwer, Moor, and
  Vandewalle]{Lathauwer:2000:BRR:354353.354405}
Lieven~De Lathauwer, Bart~De Moor, and Joos Vandewalle.
\newblock On the best rank-1 and rank-(r1,r2,. . .,rn) approximation of
  higher-order tensors.
\newblock \emph{SIAM J. Matrix Anal. Appl.}, 21\penalty0 (4):\penalty0
  1324--1342, March 2000.
\newblock ISSN 0895-4798.
\newblock \doi{10.1137/S0895479898346995}.
\newblock URL \url{http://dx.doi.org/10.1137/S0895479898346995}.

\bibitem[Lebedev et~al.(2014)Lebedev, Ganin, Rakhuba, Oseledets, and
  Lempitsky]{DBLP:journals/corr/LebedevGROL14}
Vadim Lebedev, Yaroslav Ganin, Maksim Rakhuba, Ivan~V. Oseledets, and Victor~S.
  Lempitsky.
\newblock Speeding-up convolutional neural networks using fine-tuned
  cp-decomposition.
\newblock \emph{CoRR}, abs/1412.6553, 2014.
\newblock URL \url{http://arxiv.org/abs/1412.6553}.

\bibitem[LeCun et~al.(1998)LeCun, Bottou, Bengio, and
  Haffner]{lecun1998gradient}
Yann LeCun, L{\'e}on Bottou, Yoshua Bengio, and Patrick Haffner.
\newblock Gradient-based learning applied to document recognition.
\newblock \emph{Proceedings of the IEEE}, 86\penalty0 (11):\penalty0
  2278--2324, 1998.

\bibitem[Liao et~al.(2013)Liao, McDermott, and Senior]{liao2013large}
Hank Liao, Erik McDermott, and Andrew Senior.
\newblock Large scale deep neural network acoustic modeling with
  semi-supervised training data for youtube video transcription.
\newblock In \emph{Automatic Speech Recognition and Understanding (ASRU), 2013
  IEEE Workshop on}, pages 368--373. IEEE, 2013.

\bibitem[Liu et~al.(2013)Liu, Yin, Wang, and Wang]{liu2013online}
Cheng-Lin Liu, Fei Yin, Da-Han Wang, and Qiu-Feng Wang.
\newblock Online and offline handwritten chinese character recognition:
  benchmarking on new databases.
\newblock \emph{Pattern Recognition}, 46\penalty0 (1):\penalty0 155--162, 2013.

\bibitem[Netzer et~al.(2011)Netzer, Wang, Coates, Bissacco, Wu, and
  Ng]{netzer2011reading}
Yuval Netzer, Tao Wang, Adam Coates, Alessandro Bissacco, Bo~Wu, and Andrew~Y
  Ng.
\newblock Reading digits in natural images with unsupervised feature learning.
\newblock In \emph{NIPS workshop on deep learning and unsupervised feature
  learning}, volume 2011, page~5. Granada, Spain, 2011.

\bibitem[Recht et~al.(2010)Recht, Fazel, and
  Parrilo]{Recht:2010:GMS:1958515.1958520}
Benjamin Recht, Maryam Fazel, and Pablo~A. Parrilo.
\newblock Guaranteed minimum-rank solutions of linear matrix equations via
  nuclear norm minimization.
\newblock \emph{SIAM Rev.}, 52\penalty0 (3):\penalty0 471--501, August 2010.
\newblock ISSN 0036-1445.
\newblock \doi{10.1137/070697835}.
\newblock URL \url{http://dx.doi.org/10.1137/070697835}.

\bibitem[Rigamonti et~al.(2013)Rigamonti, Sironi, Lepetit, and
  Fua]{rigamonti2013learning}
Roberto Rigamonti, Amos Sironi, Vincent Lepetit, and Pascal Fua.
\newblock Learning separable filters.
\newblock In \emph{Computer Vision and Pattern Recognition (CVPR), 2013 IEEE
  Conference on}, pages 2754--2761. IEEE, 2013.

\bibitem[Sainath et~al.(2013)Sainath, Kingsbury, Sindhwani, Arisoy, and
  Ramabhadran]{sainath2013low}
Tara~N Sainath, Brian Kingsbury, Vikas Sindhwani, Ebru Arisoy, and Bhuvana
  Ramabhadran.
\newblock Low-rank matrix factorization for deep neural network training with
  high-dimensional output targets.
\newblock In \emph{Acoustics, Speech and Signal Processing (ICASSP), 2013 IEEE
  International Conference on}, pages 6655--6659. IEEE, 2013.

\bibitem[Sermanet et~al.(2012)Sermanet, Chintala, and
  LeCun]{sermanet2012convolutional}
Pierre Sermanet, Sandhya Chintala, and Yann LeCun.
\newblock Convolutional neural networks applied to house numbers digit
  classification.
\newblock In \emph{Pattern Recognition (ICPR), 2012 21st International
  Conference on}, pages 3288--3291. IEEE, 2012.

\bibitem[Van~Loan(2000)]{van2000ubiquitous}
Charles~F Van~Loan.
\newblock The ubiquitous kronecker product.
\newblock \emph{Journal of computational and applied mathematics}, 123\penalty0
  (1):\penalty0 85--100, 2000.

\bibitem[Van~Loan and Pitsianis(1993)]{van1993approximation}
Charles~F Van~Loan and Nikos Pitsianis.
\newblock \emph{Approximation with Kronecker products}.
\newblock Springer, 1993.

\bibitem[Xue et~al.(2013)Xue, Li, and Gong]{xue2013restructuring}
Jian Xue, Jinyu Li, and Yifan Gong.
\newblock Restructuring of deep neural network acoustic models with singular
  value decomposition.
\newblock In \emph{INTERSPEECH}, pages 2365--2369, 2013.

\bibitem[Zhang et~al.(2009)Zhang, Guo, Chen, and Li]{zhang2009hcl2000}
Honggang Zhang, Jun Guo, Guang Chen, and Chunguang Li.
\newblock Hcl2000-a large-scale handwritten chinese character database for
  handwritten character recognition.
\newblock In \emph{Document Analysis and Recognition, 2009. ICDAR'09. 10th
  International Conference on}, pages 286--290. IEEE, 2009.

\bibitem[Zhang et~al.(2014{\natexlab{a}})Zhang, Zou, Ming, He, and
  Sun]{zhang2014efficient}
Xiangyu Zhang, Jianhua Zou, Xiang Ming, Kaiming He, and Jian Sun.
\newblock Efficient and accurate approximations of nonlinear convolutional
  networks.
\newblock \emph{arXiv preprint arXiv:1411.4229}, 2014{\natexlab{a}}.

\bibitem[Zhang et~al.(2014{\natexlab{b}})Zhang, Chuangsuwanich, and
  Glass]{zhang2014extracting}
Yu~Zhang, Ekapol Chuangsuwanich, and James Glass.
\newblock Extracting deep neural network bottleneck features using low-rank
  matrix factorization.
\newblock In \emph{Proc. ICASSP}, 2014{\natexlab{b}}.

\end{thebibliography}

\end{document}